\documentclass[letterpaper, 10 pt, conference]{ieeeconf}  
\IEEEoverridecommandlockouts                              
\usepackage{graphicx}
\usepackage{booktabs}
\usepackage{multirow}
\usepackage{cuted}
\usepackage{capt-of}
\usepackage{amsmath}
\usepackage{amsfonts}
\usepackage{tabularx}
\usepackage{array}
\usepackage{float}
\newcolumntype{Y}{>{\centering\arraybackslash}X}
\usepackage[table]{xcolor}
\usepackage{hyperref}
\title{\LARGE \bf
VT-WAM: Visual-Tactile World Action Model for Contact-Rich Manipulation
}

\author{Shuai Tian$^{1,2}$, Yupeng Zheng$^{1,2,3*}$, Yuhang Zheng$^{4}$, Songen Gu$^{5}$, Yujie Zang$^{3,4}$,\\
Yuxing Qin$^{1,2}$, Weize Li$^{3}$, Haoran Li$^{1,2,\dagger}$, Wenchao Ding$^{3,\dagger}$, Dongbin Zhao$^{1,2}$\\[0.15em]
\small $^{1}$ SKL-MAIS, Institute of Automation, Chinese Academy of Sciences \\
\small $^{2}$ School of Artificial Intelligence, University of Chinese Academy of Sciences \\
\small $^{3}$ TARS Robotics \quad
$^{4}$ National University of Singapore \quad
$^{5}$ Fudan University\\
\small $^{*}$ Project Leader
\small $^{\dagger}$ Corresponding Author
}

\begin{document}

\maketitle

\thispagestyle{empty}
\pagestyle{empty}

\begin{abstract}
    Contact-rich manipulation requires policies to react to local deformation, pressure, slip, and friction, yet these cues are temporally sparse and often invisible in visual observations.
    Existing visual-tactile policies usually feed tactile observations directly into action prediction, but rarely model tactile deformation dynamics during action generation.
    In this paper, we introduce VT-WAM, a Visual-Tactile World Action Model that jointly learns future visual prediction, tactile deformation prediction, and action prediction within a unified flow matching framework. 
    In particular, VT-WAM introduces (1) Asymmetric Mixture-of-Transformers (MoT) Attention to bridge a first-frame visual anchor with temporal tactile dynamics, and (2) contact-gated Action-Visual-Tactile Attention Guidance (AVTAG) to encourage action queries to rely on tactile evidence during contact phases.
    Across six real-world contact-rich manipulation tasks, VT-WAM achieves a 71.67\% average success rate, outperforming Fast-WAM by 26.67\% and OmniVTLA by 35.84\%. 
    Ablations demonstrate that modeling tactile deformation dynamics and guiding contact-phase tactile attention are both important for contact-rich tasks.
    Project Website: \href{https://vt-wam.github.io/}{https://vt-wam.github.io/}.
\end{abstract}


\section{Introduction}
Contact-rich manipulation poses a core and persistent challenge in robotic manipulation, essential for practical deployment. Unlike free-space manipulation, these tasks rely on local interaction states, including deformation, pressure, slip, and friction. These states are often weakly visible, transient, or occluded in visual observations, making vision-centric policies~\cite{chi2025diffusion, physical2025pi05} unreliable when execution demands tactile-informed adjustments.

\begin{figure}
    \centering
    \includegraphics[width=\linewidth]{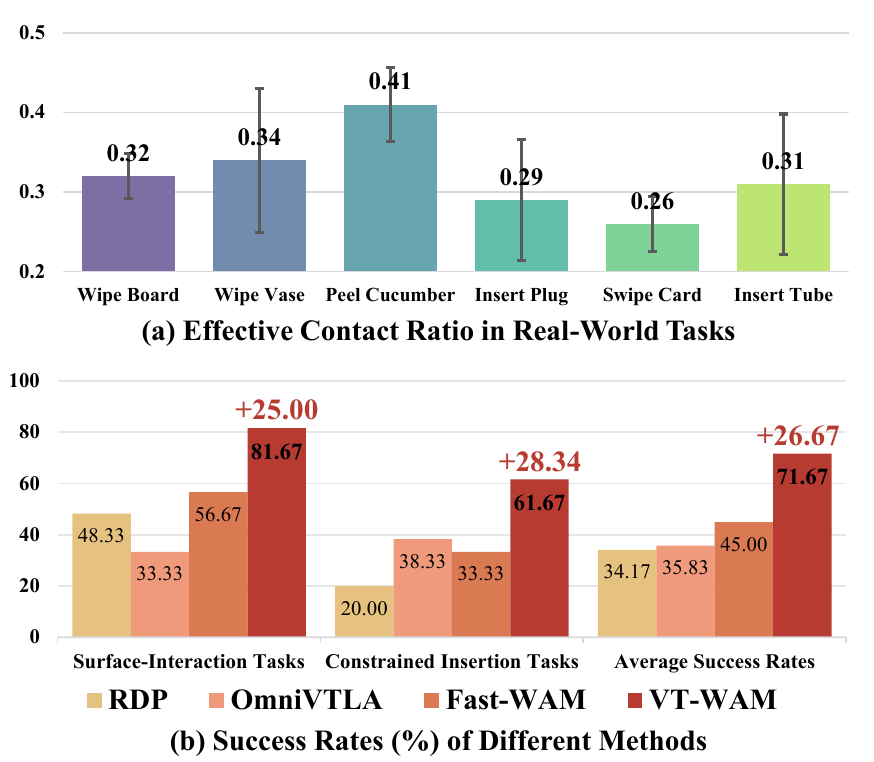}
    \caption{\textbf{Sparse tactile dynamics provide decisive evidence for contact-rich manipulation.} (a) Across six real-world tasks, tactile responses appear mainly around short contact events, making the informative signal temporally sparse. (b) By coupling action prediction with tactile deformation dynamics, VT-WAM improves the average success rate from 45.00\% with Fast-WAM to 71.67\%, with consistent gains on both surface-interaction and constrained insertion tasks.}
    \label{fig:teaser}
\end{figure}

Recent visual-tactile policies~\cite{xue2025reactive,bi2026vla,cheng2025omnivtla} have introduced tactile sensing into action prediction and have made progress on contact-rich tasks. However, these policies often fail to exploit tactile information fully~\cite{hansen2022visuotactile}. The reason is that contact-rich manipulation primarily depends on local contact evolution rather than global scene variations. Specifically, tactile deformation evolves and provides force feedback only during brief contact phases, as shown in Fig.~\ref{fig:teaser}(a). In contrast, visual observations provide dense scene-level information in most frames. This temporal imbalance in information availability causes neural networks to favor visual evidence during joint training, while tactile signals remain underutilized.

To address this issue, our key insight is to couple action prediction with tactile evolution, enabling the policy to leverage tactile changes during contact phases. Recently, World Action Models (WAMs) have provided the ability to predict world dynamics by coupling action prediction and video prediction~\cite{zhu2025unified, li2026causal, yuan2026fast}. Building on world action models, we propose VT-WAM, a visual-tactile world action model that jointly learns future visual prediction, tactile deformation prediction, and action prediction within a unified flow matching framework.

In particular, VT-WAM has two core modules that make tactile dynamics useful for action prediction. Asymmetric MoT Attention routes action tokens to a first-frame visual anchor for scene context and to the full tactile sequence for contact evolution. This enables visual-cache inference mode without discarding tactile dynamics needed for contact phases. Contact-gated AVTAG further reduces visual-dominance bias by applying a training-only hinge ranking loss that encourages action queries to attend to tactile evidence during contact phases. This auxiliary guidance makes the model rely more on tactile dynamics when contact information is physically informative, without changing the inference-time architecture.

We evaluate VT-WAM on six real-world contact-rich tasks, covering surface-interaction and constrained insertion regimes. As shown in Fig.~\ref{fig:teaser}(b), VT-WAM achieves a 71.67\% success rate and outperforms the baseline Fast-WAM~\cite{yuan2026fast} by 26.67\%. Detailed ablation studies of tactile dynamics modeling methods and attention guidance demonstrate the effectiveness of our core designs in contact-rich tasks.

Our main contributions are summarized as follows:
\begin{itemize}
    \item We formulate VT-WAM to couple tactile deformation dynamics with action prediction through joint visual-tactile-action flow matching.
    \item We introduce Asymmetric MoT Attention and AVTAG to enable a visual anchor, temporal tactile dynamics, and contact-phase tactile guidance.
    \item We validate VT-WAM on six real-world tasks, reaching 71.67\% average success, 26.67\% above Fast-WAM; ablations confirm both designs.
\end{itemize}

\section{Related Work}
\subsection{Tactile Policy for Contact-Rich Manipulation}
Tactile sensing is useful for contact-rich manipulation because it provides local interaction information. Existing tactile policies commonly use tactile signals in three ways. One line of work augments diffusion policies~\cite{chi2025diffusion} with tactile conditioning, including FARM~\cite{helmut2025tactile}, TacDiffusion~\cite{wu2025tacdiffusion}, PolyTouch~\cite{zhao2025polytouch}, and KineDex~\cite{zhang2025kinedex}. Another line develops reactive or dual-system policies that use tactile or force feedback for online contact correction, including RDP~\cite{xue2025reactive}, Force Policy~\cite{fang2026force}, and M2-ResPolicy~\cite{li2026master}. Recent Vision-Language-Action (VLA) models further incorporate tactile feedback into large-scale policy architectures such as BiTLA~\cite{yang2025bitla}, OmniVTLA~\cite{cheng2025omnivtla}, VTLA~\cite{zhang2025vtla}, Tactile-VLA~\cite{huang2025tactile}, TaF-VLA~\cite{huang2026tactile}, and VLA-Touch~\cite{bi2026vla}. Beyond using tactile observations as policy inputs, recent visual-tactile world models explicitly predict contact evolution: VT-WM jointly models visual and tactile observations in a latent recurrent state space and uses the learned dynamics for planning~\cite{higuera2026visuo}, while OmniVTA predicts visual-tactile evolution and incorporates the predicted tactile into adaptive policy fusion and reflex control~\cite{zheng2026omnivta}. These methods show that tactile deformations are valuable for contact-rich tasks, but tactile prediction is still used indirectly through planning or downstream action modules. In contrast, our VT-WAM couples tactile prediction with action prediction in a single flow matching objective with Asymmetric MoT Attention and contact-gated guidance, enabling tactile dynamics-aware action prediction. 

\subsection{World Action Models for Manipulation}
Compared with Vision-Language-Action (VLA) models that directly predict actions from current observations and language instructions, World Action Models~\cite{wang2026world} incorporate future-state prediction into action prediction. Existing WAMs are commonly organized into cascaded and joint architectures according to how future-state prediction is coupled with action prediction. Cascaded WAMs first synthesize future visual states or intermediate plans, and then derive executable actions from the predicted futures, with representative methods including UniPi~\cite{du2023learning}, VLP~\cite{du2024video}, RoboEnvision~\cite{yang2025roboenvision}, and Dream4manip~\cite{gu2026say}. Joint WAMs instead learn future dynamics and action prediction within a shared generative objective, as in Fast-WAM~\cite{yuan2026fast}, DreamZero~\cite{ye2026world}, Motus~\cite{bi2025motus}, LingBot-VA~\cite{li2026causal}, GigaWorld-Policy~\cite{ye2026gigaworld}, and UWM~\cite{zhu2025unified}. However, existing WAMs mainly model visual dynamics for action prediction. VT-WAM extends WAMs to tactile deformation dynamics, allowing contact evolution to directly inform action prediction.

\section{Method}\label{sec:method}
\begin{figure*}[t]
  \centering
  \includegraphics[width=0.98\textwidth]{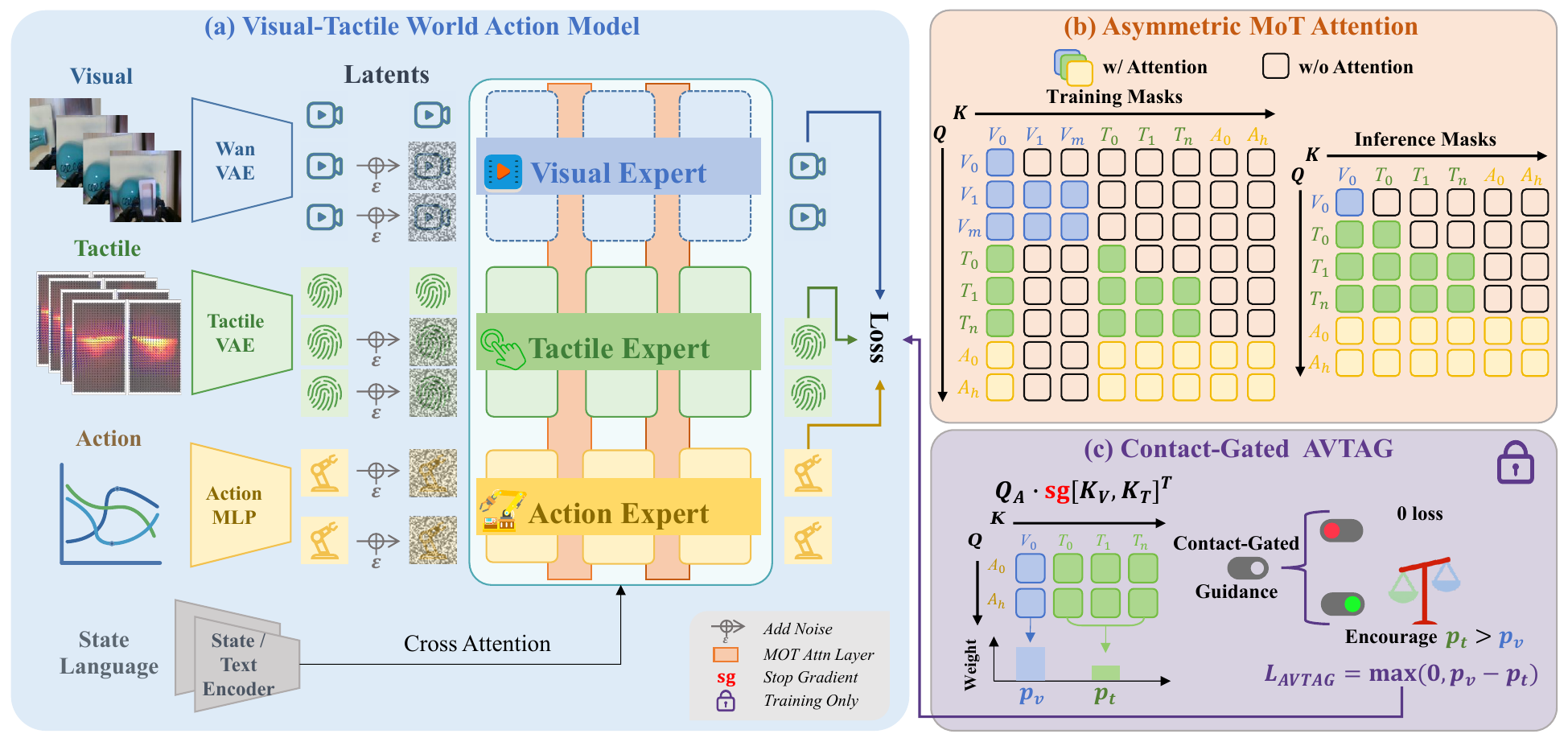}
  \caption{\textbf{Overview of VT-WAM.} (a) Joint visual-tactile-action flow matching with three modality-specific experts connected by Asymmetric MoT Attention. (b) Attention masks in Asymmetric MoT Attention during training and inference. (c) Contact-gated AVTAG applies a training-only hinge ranking loss that encourages action queries to prioritize tactile evidence during contact phases.}
  \label{fig:method_overview}
\end{figure*}

VT-WAM is a visual-tactile World Action Model for contact-rich manipulation. Given wrist camera observations $\mathbf{O}^v$, tactile deformation observations $\mathbf{O}^t$, proprioceptive state $\mathbf{s}$, language instruction $c$, and action chunk $\mathbf{A}$, VT-WAM jointly learns future visual prediction, tactile deformation prediction, and action prediction within a unified flow matching framework. Fig.~\ref{fig:method_overview} illustrates the overall VT-WAM architecture, including the visual-tactile-action expert backbone, Asymmetric MoT Attention, contact-gated AVTAG, and the training and inference procedures.

\subsection{The Architecture of VT-WAM}

As shown in Fig.~\ref{fig:method_overview}(a), VT-WAM uses a visual-tactile-action expert architecture. The visual expert encodes wrist camera tokens as global scene context, the tactile expert models local contact evolution from tactile deformation tokens, and the action expert predicts the action chunk from visual and tactile evidence. Asymmetric MoT Attention connects the three experts, enabling joint visual, tactile, and action prediction within one backbone.

VT-WAM first maps each modality into a token sequence. The wrist camera sequence $\mathbf{O}^v\in\mathbb{R}^{T_v\times 3\times H\times W}$ is encoded by the Wan2.2 video VAE~\cite{wan2025wan} and patchified into visual tokens $\mathbf{X}_v\in\mathbb{R}^{N_v\times d}$. The tactile deformation sequence $\mathbf{O}^t\in\mathbb{R}^{T_t\times 6\times H_t\times W_t}$ contains 3D deformation fields from two tactile surfaces. Following OmniVTA~\cite{zheng2026omnivta}, a pretrained tactile VAE encodes this sequence into tactile tokens $\mathbf{X}_t\in\mathbb{R}^{N_t\times d}$. Each action chunk $\mathbf{A}\in\mathbb{R}^{S_a\times D_a}$ is linearly projected into action tokens $\mathbf{X}_a\in\mathbb{R}^{S_a\times d}$. The language instruction and proprioceptive state are provided to each expert through cross-attention~\cite{vaswani2017attention}.

At the $l$-th Asymmetric MoT Attention layer, each expert computes query, key, and value tensors from its own token stream. These tensors are then concatenated in the order of visual, tactile, and action tokens:
\[
\begin{aligned}
\mathbf{Q}^{(l)} &= [\mathbf{Q}_v^{(l)};\mathbf{Q}_t^{(l)};\mathbf{Q}_a^{(l)}],\\
\mathbf{K}^{(l)} &= [\mathbf{K}_v^{(l)};\mathbf{K}_t^{(l)};\mathbf{K}_a^{(l)}],\\
\mathbf{V}^{(l)} &= [\mathbf{V}_v^{(l)};\mathbf{V}_t^{(l)};\mathbf{V}_a^{(l)}].
\end{aligned}
\]
Asymmetric MoT Attention performs masked attention over this concatenated token sequence:
\[
\begin{aligned}
\mathbf{P}^{(l)}
&=
\mathrm{Softmax}\left(
\frac{\mathbf{Q}^{(l)}(\mathbf{K}^{(l)})^\top}{\sqrt d}
+\mathbf{M}
\right),\\
\mathbf{Y}^{(l)}
&=
\mathbf{P}^{(l)}\mathbf{V}^{(l)}.
\end{aligned}
\]
Here $\mathbf{P}^{(l)}$ denotes the attention map over visual, tactile, and action tokens, and $\mathbf{M}$ determines which query tokens can attend to which key tokens. The resulting output $\mathbf{Y}^{(l)}$ gives the updated token features for the visual, tactile, and action experts. After the Asymmetric MoT Attention layers, modality-specific projection heads predict velocity fields for the visual, tactile, and action tokens under the flow matching objective. The visual and tactile experts supervise future visual prediction and tactile deformation prediction during training, while the action head produces the action chunk for control.

\subsection{Asymmetric MoT Attention}

Asymmetric MoT Attention controls how visual, tactile, and action tokens exchange information in MoT~\cite{liang2025mixture} layers. The design is motivated by two requirements of contact-rich control. First, wrist camera observations mainly provide global scene context, while tactile deformation provides the key evidence for contact interaction. Therefore, action prediction should access the tactile sequence to capture contact evolution. Second, while tactile dynamics are essential for contact-rich control, denoising future visual tokens introduces unnecessary latency during deployment. VT-WAM therefore uses an asymmetric readout: action tokens attend to the tactile sequence for contact dynamics, but attend only to the first-frame visual tokens for global context.

Fig.~\ref{fig:method_overview}(b) illustrates how this readout is implemented in training and inference. During training, VT-WAM keeps the visual, tactile, and action branches in one joint flow matching model, so future visual, tactile, and action predictions are optimized together. During inference, future visual tokens are removed. The tactile and action branches use the first-frame visual anchor, and action tokens attend to the tactile latent sequence being denoised. This preserves contact-dynamics modeling while avoiding the cost of future visual prediction.

We formalize the cross-modal mask used by the asymmetric readout below. The visual tokens are divided into the first-frame visual anchor and future visual tokens. Let $F_v$ denote the number of first-frame visual tokens, $N_v$ the number of all visual tokens, $N_t$ the number of tactile tokens, and $S_a$ the number of action tokens. VT-WAM packs these tokens in the order $[\mathbf{X}_v;\mathbf{X}_t;\mathbf{X}_a]$ and applies a blockwise attention mask $\mathbf{M}$, where rows correspond to query tokens being updated and columns correspond to key tokens available as information sources. A zero entry allows attention, while $-\infty$ blocks attention. Same-modality attention is retained within each expert, and the following rules specify the cross-modal information flow.

For the visual expert, tactile and action keys are masked out so that local contact deformation and future action tokens do not modify the visual representation:
\[
\begin{aligned}
\mathbf{M}_{v\rightarrow t}
&=
-\infty\cdot\mathbf{1}_{N_v\times N_t},\\
\mathbf{M}_{v\rightarrow a}
&=
-\infty\cdot\mathbf{1}_{N_v\times S_a}.
\end{aligned}
\]

For the tactile expert, only the first-frame visual anchor is visible among visual tokens. This grounds tactile dynamics in the global scene context while preventing dependence on future visual tokens. Action keys are also masked out in this readout mask:
\[
\begin{aligned}
\mathbf{M}_{t\rightarrow v}
&=
[\mathbf{0}_{N_t\times F_v}
\mid
-\infty\cdot\mathbf{1}_{N_t\times (N_v-F_v)}],\\
\mathbf{M}_{t\rightarrow a}
&=
-\infty\cdot\mathbf{1}_{N_t\times S_a}.
\end{aligned}
\]

For the action expert, the mask exposes the first-frame visual anchor and the full tactile sequence, matching the visual-cache inference mode used for control:
\[
\begin{aligned}
\mathbf{M}_{a\rightarrow v}
&=
[\mathbf{0}_{S_a\times F_v}
\mid
-\infty\cdot\mathbf{1}_{S_a\times (N_v-F_v)}],\\
\mathbf{M}_{a\rightarrow t}
&=
\mathbf{0}_{S_a\times N_t}.
\end{aligned}
\]
Thus, Asymmetric MoT Attention keeps visual representations stable, grounds tactile dynamics in visual context, and provides action prediction with both visual anchor and contact-evolution information.

\subsection{Contact-Gated Action-Visual-Tactile Attention Guidance}

Although Asymmetric MoT Attention allows action tokens to attend to tactile tokens, joint training can still favor visual evidence over tactile evidence. This is because visual and tactile signals are imbalanced in contact-rich tasks. Visual observations provide dense scene-level information across most frames. In contrast, tactile deformation is local and temporally sparse: it becomes informative mainly during short contact intervals and remains weak or inactive outside contact~\cite{zheng2026omnivta}. As a result, the joint flow matching objective can reduce training loss by relying primarily on visual context, while underusing tactile dynamics that are critical during contact phases.

To mitigate this imbalance, VT-WAM introduces Action-Visual-Tactile Attention Guidance (AVTAG), illustrated in Fig.~\ref{fig:method_overview}(c). AVTAG adds a training-only auxiliary attention objective that computes the relative attention from action queries to visual and tactile evidence. During contact phases, it applies a contact-gated hinge ranking loss that penalizes cases where relative tactile attention is lower than relative visual attention. This guides action queries to increase tactile attention when local physical interaction is informative.

AVTAG constructs an auxiliary attention distribution from action queries to visual and tactile keys. For clarity, we omit the layer index and use $\mathbf{Q}_a$, $\mathbf{K}_v$, and $\mathbf{K}_t$ to denote the action queries and visual-tactile keys from MoT Attention layers. Let $\mathbf{K}_{\mathrm{vt}}=[\mathbf{K}_v;\mathbf{K}_t]$ denote the concatenated visual and tactile keys. To guide action queries without directly changing the visual and tactile key representations, we apply stop-gradient to $\mathbf{K}_{\mathrm{vt}}$ and define
\[
\mathbf{P}_{\mathrm{vt}}
=
\mathrm{Softmax}\left(
\frac{\mathbf{Q}_a\mathrm{sg}(\mathbf{K}_{\mathrm{vt}})^\top}{\sqrt d}
\right),
\]
where $\mathrm{sg}(\cdot)$ denotes stop-gradient. This auxiliary attention is used only for the AVTAG loss, so its gradients guide the action queries while leaving the visual and tactile keys optimized by the main flow matching objective.

For each action token $r\in\{1,\ldots,S_a\}$, AVTAG sums the auxiliary attention assigned to visual keys and tactile keys:
\[
\alpha_v(r)=\sum_{j\in\mathrm{visual}}\mathbf{P}_{\mathrm{vt}}[r,j],
\qquad
\alpha_t(r)=\sum_{j\in\mathrm{tactile}}\mathbf{P}_{\mathrm{vt}}[r,j].
\]
These two quantities are then normalized into relative visual and tactile attention weights:
\[
p_v(r)=\frac{\alpha_v(r)}{\alpha_v(r)+\alpha_t(r)},
\qquad
p_t(r)=\frac{\alpha_t(r)}{\alpha_v(r)+\alpha_t(r)}.
\]

AVTAG applies this guidance only to action tokens in contact phases. Let $\mathcal{C}$ denote contact-phase action tokens, identified by pronounced tactile deformation. The auxiliary loss is defined as
\[
L_{\mathrm{AVTAG}}
=
\mathbb{E}_{r\in\mathcal{C}}
\left[
\max\left(0,\; p_v(r)-p_t(r)\right)
\right].
\]
This hinge ranking loss penalizes visual-dominant attention during contact phases, and incurs no penalty once $p_t(r)\ge p_v(r)$.

\begin{figure}
    \centering
    \includegraphics[width=\linewidth]{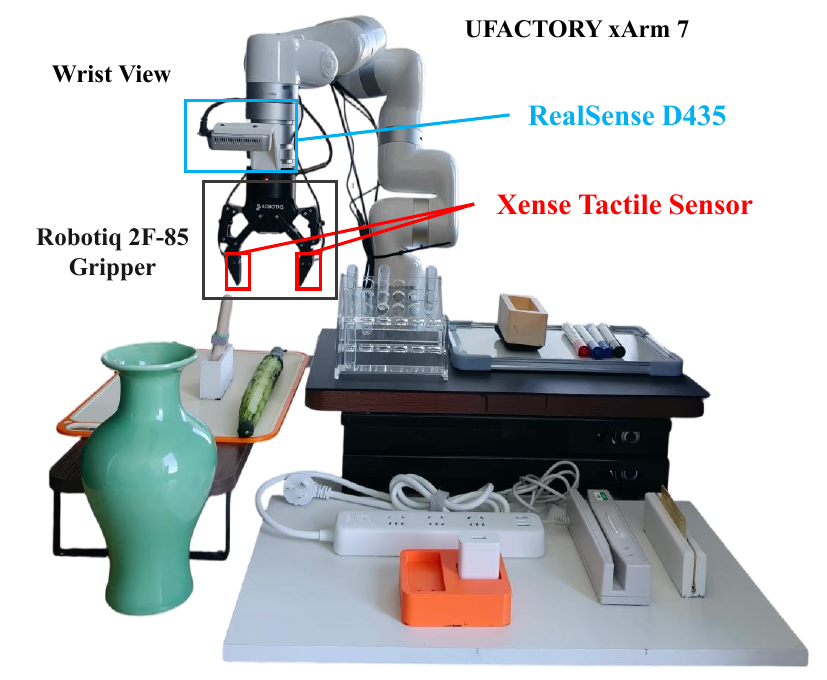}
    \caption{\textbf{Real-world experimental platform.} The setup uses a 7-DoF xArm7 robot with a Robotiq 2F-85 gripper, a wrist camera, and paired gripper-mounted Xense tactile sensors. The scene includes the representative objects used in our experiments.}
    \label{fig:experiment_setup}
\end{figure}

\subsection{Training Objective and Efficient Inference}

\subsubsection{Flow Matching Training Objective}

VT-WAM is trained with a joint flow matching objective over visual, tactile, and action tokens. The visual, tactile, and action experts each predict the velocity field of their corresponding modality, yielding
\[
L_{\mathrm{Flow}}
=
\lambda_v L_v+\lambda_t L_t+\lambda_a L_a,
\]
\[
L_v=\mathbb{E}\|\hat{\mathbf{f}}^v-\mathbf{f}_v^*\|^2,\quad
L_t=\mathbb{E}\|\hat{\mathbf{f}}^t-\mathbf{f}_t^*\|^2,\quad
L_a=\mathbb{E}\|\hat{\mathbf{f}}^a-\mathbf{f}_a^*\|^2.
\]
Here $\hat{\mathbf{f}}^v$, $\hat{\mathbf{f}}^t$, and $\hat{\mathbf{f}}^a$ denote the predicted velocity fields for visual, tactile, and action tokens, and $\mathbf{f}_v^*$, $\mathbf{f}_t^*$, and $\mathbf{f}_a^*$ denote the corresponding flow matching targets. When AVTAG is enabled, the full training objective is
\[
L_{\mathrm{Train}}
=
L_{\mathrm{Flow}}
+
\lambda_{\mathrm{AVTAG}}L_{\mathrm{AVTAG}}.
\]

\begin{figure*}
    \centering
    \includegraphics[width=1.0\textwidth]{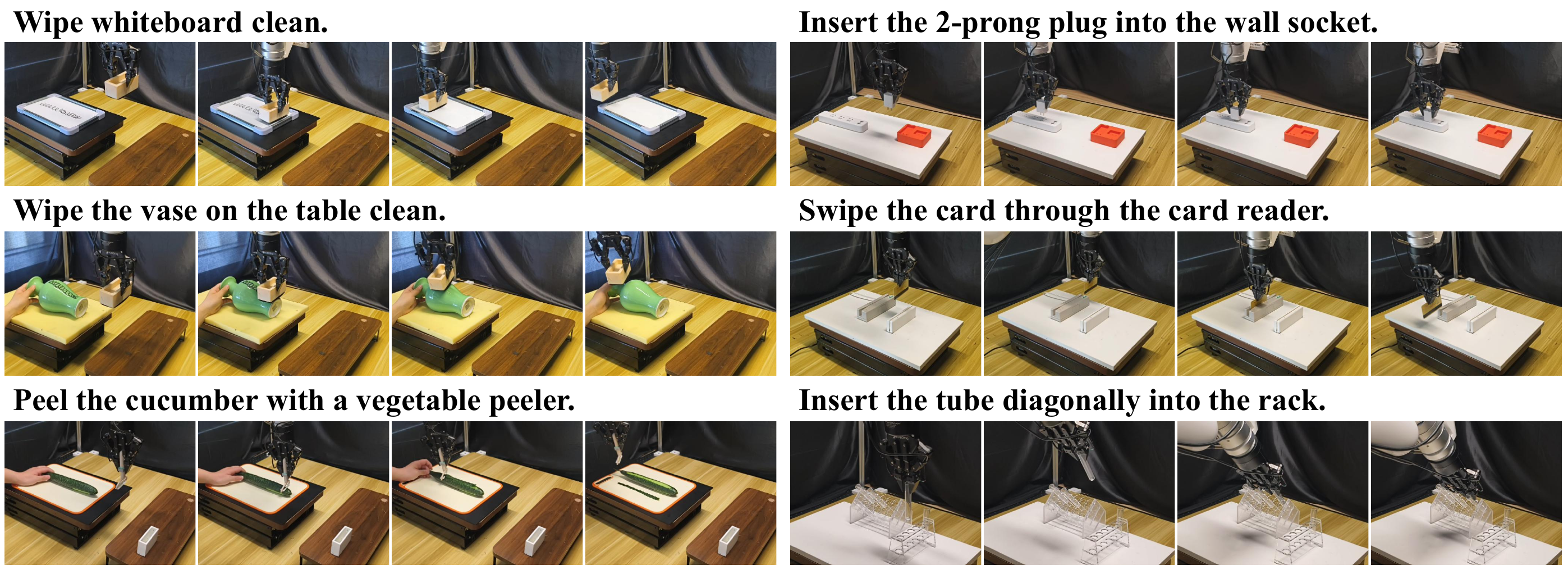}
    \caption{\textbf{Overview of real-world contact-rich manipulation tasks.} We evaluate VT-WAM on six real-world tasks covering two interaction regimes: surface-interaction tasks and constrained insertion tasks.}
    \label{fig:task_progress}
    \vspace{-10pt}
\end{figure*}

\begin{table*}
\centering
\caption{\textbf{Success rates on real-world contact-rich tasks.}}
\label{tab:success_rate}
\normalsize
\resizebox{\textwidth}{!}{
\begin{tabular}{lccccccccc}
\toprule
\multirow{2}{*}{\textbf{Method}} & \multicolumn{4}{c}{\textbf{Surface-Interaction Tasks}} & \multicolumn{4}{c}{\textbf{Constrained Insertion Tasks}} & \multirow{2}{*}{\textbf{Average}} \\
\cmidrule(lr){2-5} \cmidrule(lr){6-9}
 & \textbf{Wipe Board} & \textbf{Wipe Vase} & \textbf{Peel Cucumber} & \textbf{Avg.} & \textbf{Insert Plug} & \textbf{Swipe Card} & \textbf{Insert Tube} & \textbf{Avg.} &  \\
\midrule
DP + Tactile~\cite{helmut2025tactile}  & 30\% & 20\% & 25\% & 25.00\% & 5\%  & 35\% & 15\% & 18.33\% & 21.67\% \\
RDP~\cite{xue2025reactive}              & 45\% & 60\% & 40\% & 48.33\% & 15\% & 35\% & 10\% & 20.00\% & 34.17\% \\
$\pi_{0.5}$~\cite{physical2025pi05}      & 40\% & 35\% & 35\% & 36.67\% & 30\% & 45\% & 10\% & 28.33\% & 32.50\% \\
OmniVTLA~\cite{cheng2025omnivtla}  & 45\% & 30\% & 25\% & 33.33\% & 40\% & 35\% & 40\% & 38.33\% & 35.83\% \\
Fast-WAM~\cite{yuan2026fast}          & 70\% & 55\% & 45\% & 56.67\% & 20\% & 55\% & 25\% & 33.33\% & 45.00\% \\
\midrule \rowcolor[HTML]{ECF4FF}
\textbf{VT-WAM} & \textbf{90\%} & \textbf{85\%} & \textbf{70\%} & \textbf{81.67\%} & \textbf{60\%} & \textbf{70\%} & \textbf{55\%} & \textbf{61.67\%} & \textbf{71.67\%} \\
\bottomrule
\end{tabular}
} 
\end{table*}

\subsubsection{Efficient Visual-Cache Inference}

VT-WAM supports two inference modes: joint inference mode and visual-cache inference mode. For visual-tactile prediction analysis, we use joint inference mode, where the model denoises visual, tactile, and action tokens together to evaluate its predictive ability. For real-world control, we use visual-cache inference mode, where the current visual observation is kept as a first-frame anchor and future visual prediction is removed. In this mode, VT-WAM denoises only tactile and action latents through Asymmetric MoT Attention: the tactile expert predicts future tactile deformation as contact evolution, and the action expert predicts the action chunk by attending to both the visual anchor and the tactile sequence. This avoids the cost of predicting future visual tokens during deployment.

\section{Experiments}

In this section, we first describe the experimental setup, including the robotic platform, implementation details, baselines, benchmark tasks, and evaluation metrics. We then evaluate VT-WAM on six contact-rich manipulation tasks, analyze visual-tactile prediction quality, and conduct ablation studies to quantify the contribution of key components.

\subsection{Experimental Setup}

\subsubsection{Robotic Platform}
To evaluate VT-WAM on real-world contact-rich manipulation tasks, we use the physical robotic platform shown in Fig.~\ref{fig:experiment_setup}. The platform consists of a 7-DoF xArm7 robot equipped with a Robotiq 2F-85 parallel gripper, a wrist camera, and two Xense tactile sensors mounted on the inner surfaces of the gripper fingers. The wrist camera captures $128 \times 128$ RGB observations at 30\,Hz. Each tactile sensor records a $35 \times 20$ three-dimensional deformation field over the contact surface at 30\,Hz.

\subsubsection{Implementation Details}
Training data are collected through human kinesthetic teaching, with 100 expert trajectories for each task. The visual, tactile, proprioceptive, and action streams are synchronized and resampled to 30\,Hz before training. VT-WAM uses pretrained Wan2.2-5B~\cite{wan2025wan} as the visual backbone and uses 1B-scale DiT models for the tactile and action experts. The loss weights are set to $\lambda_v=\lambda_t=\lambda_a=1$ and $\lambda_{\mathrm{AVTAG}}=0.05$ in all experiments. We optimize models with AdamW using a learning rate of $1 \times 10^{-4}$, weight decay of $1 \times 10^{-2}$, bf16 mixed precision, gradient clipping at 1.0, and cosine learning-rate decay after a $5\%$ warmup. Training is conducted on NVIDIA A100 (80GB) GPUs. During inference evaluation, VT-WAM runs on a remote NVIDIA A100 inference server and uses 10 denoising steps for action prediction.

\begin{figure*}
    \centering
    \includegraphics[width=1.0\textwidth]{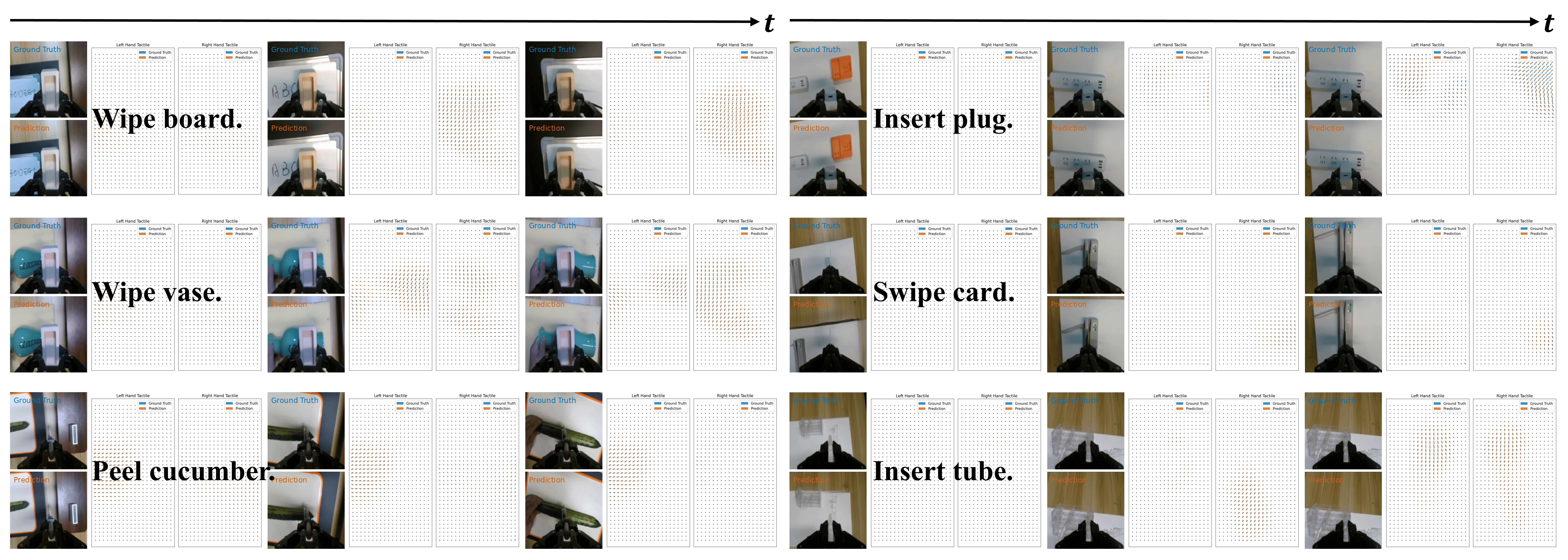}
    \vspace{-20pt}
    \caption{\textbf{Visual-Tactile Prediction Results across Six Tasks.} For visualization, VT-WAM predicts wrist camera observations together with tactile deformation fields. Blue denotes ground truth, and orange indicates prediction.}
    \label{fig:vt_prediction}
    \vspace{-10pt}
\end{figure*}

\subsubsection{Baselines}

We compare VT-WAM with representative baselines that cover visuomotor, VLA, and WAM policies:
\begin{itemize}
    \item \textbf{DP + Tactile}~\cite{helmut2025tactile}: a tactile-conditioned diffusion policy that predicts action chunks from robot states, wrist camera, and tactile observations.
    \item \textbf{RDP}~\cite{xue2025reactive}: a reactive visual-tactile policy that uses tactile feedback for online action refinement.
    \item \textbf{\boldmath$\pi_{0.5}$}~\cite{physical2025pi05}: a general vision-language-action policy without tactile input.
    \item \textbf{OmniVTLA}~\cite{cheng2025omnivtla}: a tactile-augmented VLA model that uses tactile observations for action prediction.
    \item \textbf{Fast-WAM}~\cite{yuan2026fast}: a world action model that models visual dynamics and predicts actions without tactile input.
\end{itemize}

All methods are trained separately on each task with the same demonstrations and evaluated on the same robot platform, task definitions, and metrics.

\subsection{Benchmark Tasks and Evaluation Metrics}

\subsubsection{Benchmark Tasks}
As shown in Fig.~\ref{fig:task_progress}, we evaluate VT-WAM on six contact-rich tasks~\cite{xue2025reactive,zheng2026omnivta} grouped into two regimes: surface-interaction tasks and constrained insertion tasks. Surface-interaction tasks include \textit{wipe board}, \textit{wipe vase}, and \textit{peel cucumber}, which require sustained motion over planar, curved, or deformable surfaces. Constrained insertion tasks include \textit{insert plug}, \textit{swipe card}, and \textit{insert tube}, which require fine alignment under tight geometric constraints or visual occlusion.

\subsubsection{Evaluation Metrics}
For each method and each task, we conduct 20 independent trials and report the success rate. For surface-interaction tasks, the score is defined in $\{0,0.5,1\}$: 0 indicates failure, 0.5 indicates completing more than half of the target region, and 1 indicates completing the full target region. For constrained insertion tasks, the score is binary in $\{0,1\}$, where 1 indicates that the object reaches the target position and 0 otherwise.

\subsection{Main Results}

\subsubsection{Task Performance}
Table~\ref{tab:success_rate} reports performance across the six contact-rich tasks. VT-WAM achieves the highest success rate among all evaluated methods. Compared with the strongest baseline Fast-WAM~\cite{yuan2026fast}, VT-WAM improves the success rate from 45.00\% to 71.67\%, corresponding to an absolute gain of 26.67\%.

In surface-interaction tasks, the wrist camera often changes only subtly during execution, while contact changes occur at the local interaction surface. Therefore, vision-based policies remain limited on these tasks: $\pi_{0.5}$ achieves 36.67\% success rate, while OmniVTLA achieves only 33.33\% despite using tactile input. This comparison suggests that using tactile observations only as policy inputs is insufficient to model tactile interaction dynamics, which may explain why it does not improve the success rate. Fast-WAM improves the success rate to 56.67\% by modeling action-conditioned visual dynamics, highlighting the benefit of action-conditioned dynamics modeling. However, Fast-WAM remains vision-only and cannot directly capture local tactile interaction. VT-WAM improves the success rate to 81.67\% by modeling tactile deformation as interaction dynamics.

In constrained insertion tasks, success depends on fine alignment rather than sustained surface coverage. Across insert plug, swipe card, and insert tube, VT-WAM achieves 61.67\% success rate, compared with 38.33\% for OmniVTLA and 33.33\% for Fast-WAM. These results suggest that tactile dynamics are also useful when the robot must correct small pose errors under tight geometric constraints. The improvement is especially clear on the insert tube, where the transparent tube makes visual alignment unreliable and successful execution requires contact-informed correction. Together with the surface-interaction results, this shows that coupling tactile deformation dynamics with action prediction improves success rates on contact-rich tasks.

\subsubsection{Visual-Tactile Prediction Results}
We analyze visual-tactile prediction results to evaluate the predictive modeling ability of VT-WAM. For this analysis, we run VT-WAM in joint inference mode to predict wrist camera observations and tactile deformation fields together; real-world control instead uses the visual-cache inference mode described in Sec.~\ref{sec:method}. Fig.~\ref{fig:vt_prediction} shows that VT-WAM predicts temporally coherent wrist camera observations and tactile deformation trajectories that capture local contact patterns, including pressure concentration and contact migration. Following OmniVTA~\cite{zheng2026omnivta}, we quantify tactile prediction quality using deformation magnitude error and directional consistency. The $l_2$ distance is computed over the full 3D deformation field, while cosine similarity is computed over non-zero deformation regions. All methods are evaluated on the same task demonstrations. As reported in Table~\ref{tab:metrics_evaluation}, VT-WAM achieves lower deformation error and higher directional consistency than the baseline models, indicating that the tactile expert learns meaningful contact deformation dynamics.

\subsection{Ablation Studies}

\begin{table}
\centering
\caption{\textbf{Tactile deformation prediction quality.}}
\label{tab:metrics_evaluation}
\setlength{\tabcolsep}{25pt}
\normalsize
\begin{tabular}{lcc}
\toprule
\textbf{Method} & \textbf{$l_2 \downarrow$} & \textbf{$\cos \uparrow$} \\
\midrule
exUMI~\cite{xu2025exumi} & 0.091 & 0.618 \\
UVA~\cite{li2025unified}   & 0.083 & 0.667 \\
\midrule \rowcolor[HTML]{ECF4FF}
\textbf{VT-WAM} & \textbf{0.077} & \textbf{0.749} \\
\bottomrule
\vspace{-10pt}
\end{tabular}
\end{table}

\begin{table}
\centering
\small 
\caption{\textbf{Ablation Study on Tactile Dynamics Modeling and Attention Guidance.}}
\label{tab:combined_ablations}
\setlength{\tabcolsep}{2pt} 
\renewcommand{\arraystretch}{1.0} 

\begin{tabularx}{\columnwidth}{clcc}
\toprule
\textbf{Models} & \textbf{Description} & \textbf{Wipe Vase} & \textbf{Insert Tube} \\
\midrule
$M_0$ & Fast-WAM~\cite{yuan2026fast} & 55\% & 25\% \\
$M_1$ & $M_0$ + Sym. ($T$ Seq.)  & 65\% & 40\% \\
$M_2$ & $M_0$ + Asym. ($T_0$)    & 40\% & 30\% \\
$M_3$ & $M_0$ + Asym. ($T$ Seq.) & 70\% & 50\% \\
\rowcolor[HTML]{ECF4FF} 
\textbf{$M_4$} & \textbf{VT-WAM: $M_3$ + AVTAG} & \textbf{85\%} & \textbf{55\%} \\
\bottomrule

\midrule[0pt]
\multicolumn{4}{@{}p{\columnwidth}@{}}{
    \scriptsize \textbf{Notes:} All variants are built from Fast-WAM by adding different tactile modeling or attention designs. \textbf{+ Sym. ($T$ Seq.)} adds tactile sequence prediction with symmetric MoT attention. \textbf{+ Asym. ($T_0$)} uses Asymmetric MoT Attention but restricts action queries to the first tactile frame. \textbf{+ Asym. ($T$ Seq.)} allows action queries to attend to the full tactile sequence without AVTAG. \textbf{VT-WAM} is the full model.
} \\
\vspace{-20pt}
\end{tabularx}
\end{table}

We conduct ablations on the wipe vase and insert tube, which represent the two benchmark regimes. Table~\ref{tab:combined_ablations} evaluates two design questions: how to incorporate tactile dynamics into action prediction, and whether contact-gated AVTAG improves the real-world success rate by guiding action queries to attend to tactile dynamics.

\textbf{Ablation 1: How to incorporate tactile dynamics into action prediction?} Compared with the visual-only Fast-WAM baseline, + Sym. ($T$ Seq.) adds tactile sequence prediction with symmetric MoT attention and improves the success rate from 55\% to 65\% on wipe vase and from 25\% to 40\% on insert tube. This result shows that introducing tactile dynamics provides additional information beyond visual dynamics. However, symmetric fusion requires future visual and tactile prediction during inference, which increases the computational cost.
To avoid future visual prediction, VT-WAM adopts Asymmetric MoT Attention, where action queries attend to the first-frame visual anchor and the full tactile sequence. The importance of temporal tactile information is shown by the comparison between + Asym. ($T_0$) and + Asym. ($T$ Seq.). When action queries are restricted to the first tactile frame, + Asym. ($T_0$) achieves only 40\% on wipe vase and 30\% on insert tube. Allowing action queries to attend to the full tactile sequence improves the success rate to 70\% and 50\%, respectively. This comparison indicates that action prediction benefits from tactile evolution over time, rather than only the initial tactile state.

\begin{figure}[t]
    \centering
    \includegraphics[width=\columnwidth]{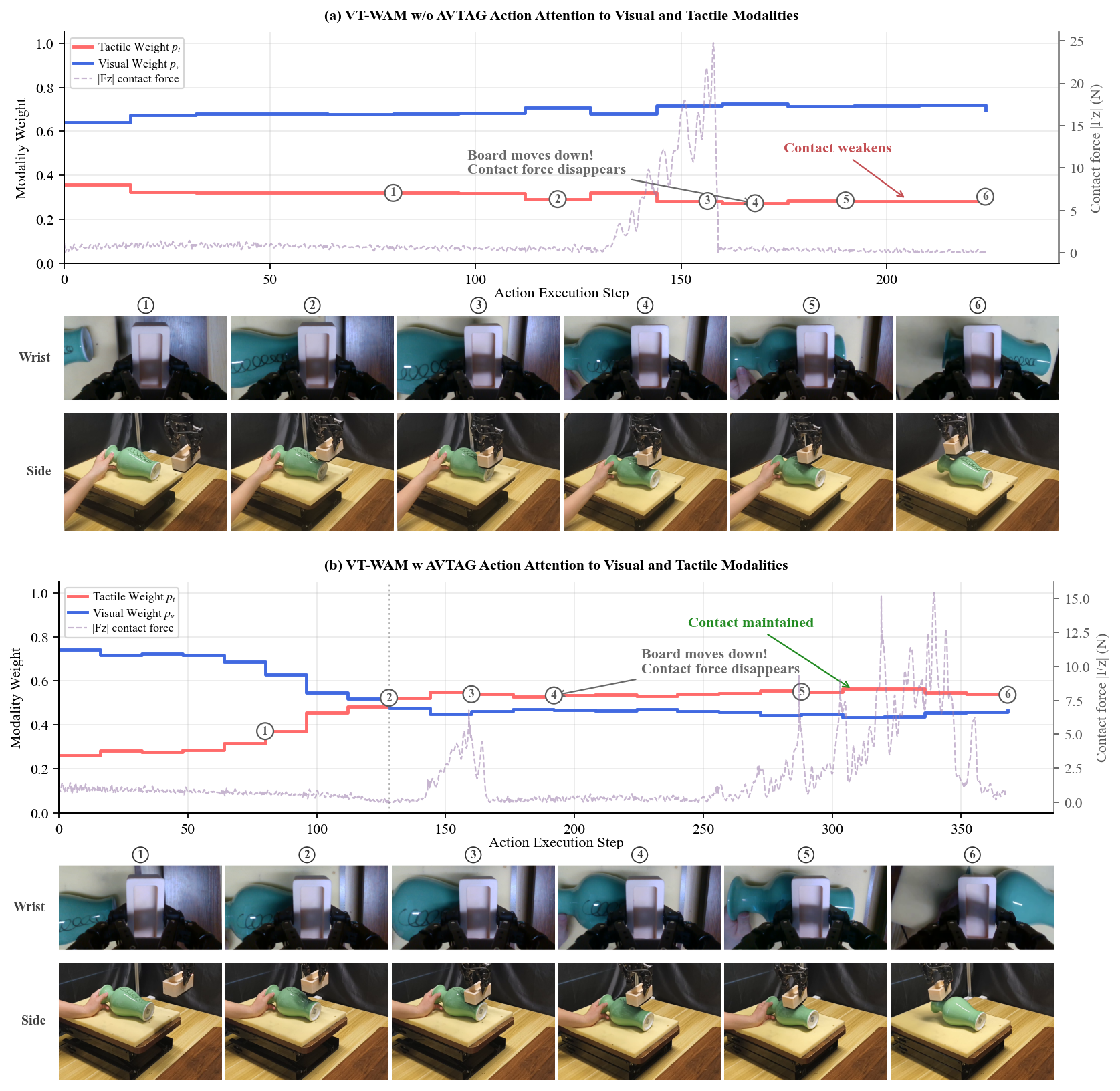}
    \vspace{-20pt}
    \caption{\textbf{AVTAG promotes tactile attention for contact recovery during vase wiping.} The red and blue curves denote relative tactile and visual attention weights $p_t$ and $p_v$ from the action expert, and the dashed curve denotes the contact force $|F_z|$ for visualization only. The wrist camera view is the only visual input available to the policy, while the side view is shown only for visualization. When the supporting plane moves downward, the wrist camera view changes only subtly and provides limited evidence about the contact phase. AVTAG encourages the action expert to attend more strongly to tactile evidence during this contact phase, enabling the model to re-establish contact and complete the task.}
    \label{fig:avtag_vt_trace}
    \vspace{-10pt}
\end{figure}

\textbf{Ablation 2: The effectiveness of the contact-gated attention guidance.} The final comparison isolates the contribution of AVTAG. The variant + Asym. ($T$ Seq.) and the full VT-WAM use the same Asymmetric MoT Attention. The only difference is the contact-gated AVTAG applied during VT-WAM training. This guidance improves the success rate from 70\% to 85\% on wipe vase and from 50\% to 55\% on insert tube, indicating that the gain comes from better contact-aware tactile attention learned during training.
Fig.~\ref{fig:avtag_vt_trace} further explains this effect through a vase-wiping trial with contact disturbance. During execution, the supporting plane of the vase is moved downward, which breaks contact between the wiping board and the vase surface. The wrist camera view is the only visual input available to the policy, whereas the side view is shown only to visualize the interaction process. After contact is lost, the wrist camera view changes only subtly, so the contact phase cannot be reliably identified from the wrist camera view alone. Without AVTAG, the action expert exhibits a nearly static attention pattern and remains dominated by visual tokens throughout the trial. This visual bias prevents the policy from responding to the contact loss. With AVTAG, the action expert increases tactile attention during the contact phase, enabling the policy to use tactile evidence to re-establish contact with the vase surface and complete the wiping task.

\section{Conclusion}

In this paper, we introduce VT-WAM, a Visual-Tactile World Action Model for contact-rich manipulation. VT-WAM extends the world action model by learning tactile deformation as temporal interaction dynamics together with action prediction, rather than using tactile observations only as auxiliary policy inputs. Real-world experiments across surface-interaction and constrained insertion tasks show that VT-WAM consistently improves over visual-only and tactile-input baselines, while tactile prediction analysis and ablations further support the importance of tactile dynamics modeling and contact-phase tactile use. These results indicate that modeling tactile deformation as interaction dynamics provides an effective way to improve action prediction in contact-rich tasks. While this work focuses on individual-task specialization for precise tactile modeling, multi-task training remains unexplored. Future research on multi-task training and scaling laws is a promising direction.



\bibliographystyle{IEEEtran}
\bibliography{sections/refs/main}

\end{document}